\documentclass[10pt,twocolumn,letterpaper]{article}

\usepackage{wacv}      
\usepackage{amsmath}
\usepackage{amssymb}
\usepackage{tabularx}
\usepackage{booktabs}
\usepackage{multirow}
\usepackage{makecell}
\usepackage{array}
\usepackage{pifont}

\newcolumntype{L}[1]{>{\raggedright\arraybackslash}p{#1}}
\newcolumntype{C}[1]{>{\centering\arraybackslash}p{#1}}
\newcolumntype{R}[1]{>{\raggedleft\arraybackslash}p{#1}}

\definecolor{wacvblue}{rgb}{0.21,0.49,0.74}
\usepackage[pagebackref,breaklinks,colorlinks,allcolors=wacvblue]{hyperref}

\def\confName{WACV}
\def\confYear{2027}

\newcounter{RZNumberOfComments}
\stepcounter{RZNumberOfComments}

\newcounter{AMNumberOfComments}
\stepcounter{AMNumberOfComments}

\definecolor{darkgreen}{RGB}{0,120,80}
\newcounter{DLNumberOfComments}

\title{ViBMR: A  Large-Scale Multi-Sequence Brain MRI Dataset and Comprehensive Evaluation for Vietnamese Radiology Report Generation}

\author{Dai Lam Le\textsuperscript{1},
Minh Khoi Nguyen\textsuperscript{1},
Son Dinh Nguyen\textsuperscript{1},
Amir Reza Jafari\textsuperscript{2},\\
Viet Ha\textsuperscript{3},
Nguyen Son\textsuperscript{4},
Reza Farahbakhsh\textsuperscript{2},
Noel Crespi\textsuperscript{2},
Phi Le Nguyen\textsuperscript{1}\\
\textsuperscript{1}AI4LIFE, Hanoi University of Science and Technology, Vietnam\\
\textsuperscript{2}SAMOVAR, Télécom SudParis, Institut Polytechnique de Paris, France\\
\textsuperscript{3}Concordia International School Hanoi, Vietnam\\
\textsuperscript{4}Hanoi – Amsterdam High School for the Gifted, Vietnam\\
{\tt\small \textsuperscript{1}\{lam.ld225982@sis, khoi.nm226050@sis, son.dn225997@sis, lenp@soict\}.hust.edu.vn}\\ 
{\tt\small \textsuperscript{2}\{amir-reza.jafari\_tehrani, reza.farahbakhsh, noel.crespi\}@telecom-sudparis.eu}\\
{\tt\small \textsuperscript{3}vietha.icloud@gmail.com, \textsuperscript{4}sonn342@os.viettel.com.vn}\\
}

\begin{document}
\maketitle

\begin{abstract}
Automated radiology report generation for multi-sequence brain MRI holds strong promise for alleviating radiologists’ workload, yet it remains a demanding problem: brain MRI studies are inherently volumetric, each pulse sequence carries complementary diagnostic information, and annotated data are critically scarce for low-resource clinical languages. Existing brain MRI-report resources are still dominated by English reports, translated clinical text, or datasets adapted from task-specific imaging benchmarks, leaving routine low-resource clinical reporting and heterogeneous multi-sequence MRI protocols insufficiently represented. To bridge this gap, we introduce \textbf{ViBMR}, to the best of our knowledge, the first large-scale multi-sequence brain MRI dataset paired with routine Vietnamese clinical reports, comprising \textbf{3{,}649 patients} and \textbf{11{,}875 MRI series} across T1, T2, FLAIR, and contrast-enhanced T1 sequences. 
Building on ViBMR, we establish a comprehensive evaluation setup for brain MRI report generation, covering representative medical vision-language models, multi-sequence inputs, and feature aggregation strategies. The results provide key insights into dataset characteristics and model behavior, while demonstrating the necessity of native low-resource clinical datasets for reliable Vietnamese-origin brain MRI report generation. Our dataset and baseline models will be released for non-commercial research.

\end{abstract}

\section{Introduction}
\begin{figure}[t]
    \centering
    \includegraphics[width=0.9\linewidth]{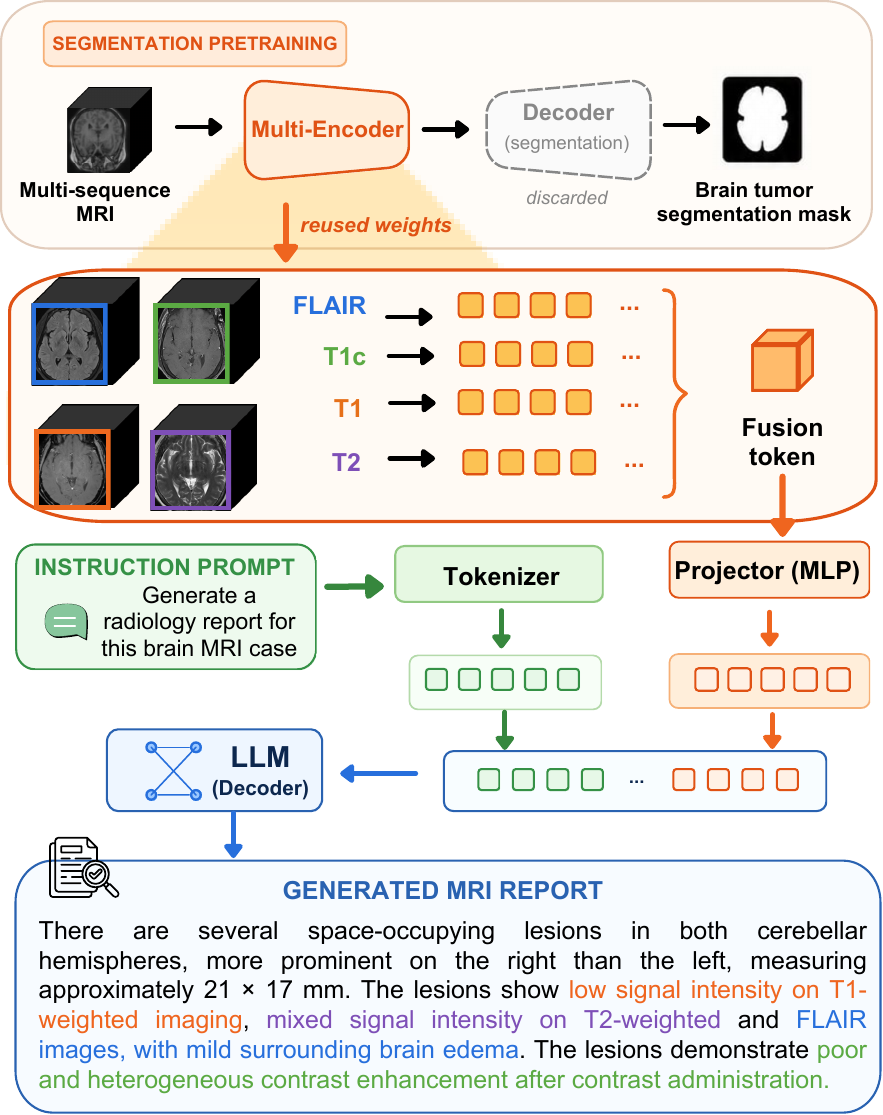}
    \caption{Overview of the proposed brain MRI report generation framework. A multi-sequence MRI encoder is pretrained on segmentation and its weights are reused to extract visual tokens from FLAIR, T1c, T1, and T2 sequences. The fused visual tokens are projected and concatenated with instruction prompt tokens, then fed into an LLM decoder to generate the final MRI report.}
    \label{fig:intro}
    \vspace{-0.4cm}
\end{figure}

Recent advances in Vision-Language Models (VLMs) have driven remarkable progress in
healthcare AI, particularly for automated radiology report generation
\cite{Chen2020R2Gen,Chen2021CMN,Tanida2023RGRG,Huang2023KiUT,Wang2023METransformer}. Among imaging modalities, brain Magnetic Resonance
Imaging (MRI) plays a pivotal role in detecting and characterizing intracranial
abnormalities. Unlike many other modalities, routine brain MRI is acquired as a
multi-sequence study, in which T1, T2, FLAIR, and contrast-enhanced T1 (T1c) provide
mutually complementary information: T2 and FLAIR are sensitive to edema and white
matter lesions, whereas T1c better characterizes enhancement patterns
\cite{Wardlaw2013STRIVE,Duering2023STRIVE2,Bakshi2001FLAIR,Smirniotopoulos2007Contrast,Lee2016CEFLAIR}. Automating the joint interpretation of these sequences
and synthesizing them into a textual report could substantially reduce radiologists'
workload and shorten diagnostic turnaround.

Despite this progress, most existing brain MRI-report datasets are constructed and
released in English, leaving low-resource clinical languages largely unaddressed
\cite{Johnson2019MIMICCXR,Forithmus2026MRRATE,Lei2024AutoRG,Bai2024M3D,LeGuellec2026PARROT,Salme2025LowResourceRRG,Nguyen2025ViMedPET}. This is a critical limitation, because radiology
reports are strongly language, institution, and population-dependent: writing style,
terminology, abbreviation usage, and the granularity of anatomical description can
differ markedly across hospitals and regions \cite{Hall2000RadiologyLanguage,Pons2016NLP,Casey2021RadiologyNLP,Nobel2022StructuredReporting}.
Furthermore, the epidemiological and morphological characteristics of Asian
populations often differ from those of European and North American cohorts in both
disease prevalence and body habitus \cite{Kim2014ICAS,Mok2014RaceSVD,Guan2022DomainAdaptation}. Together with differences in language, reporting conventions, acquisition protocols, and dataset composition, these factors motivate our cross-dataset analysis in \hyperref[rq:dataset]{RQ1}.
Beyond the language barrier, some existing ``report generation'' datasets do not
originate from routine clinical reports. Instead, they are repurposed from datasets
built for other tasks such as lesion segmentation, with findings and impressions
subsequently authored by hired radiologists for pre-annotated regions
\cite{Menze2015BraTS,Bakas2018BraTS,Lei2024AutoRG}. This construction process yields report structures that
deviate from the actual hospital workflow, in which reports are formed directly from
a radiologist's reading of the scan.

Addressing these challenges requires two ingredients in tandem: (i) a dataset for a
low-resource language that faithfully reflects routine clinical reporting, and
(ii) a comprehensive empirical study that clarifies which visual components and fusion
strategies are genuinely beneficial for multi-sequence brain MRI report generation.
Existing resources, however, satisfy neither requirement, especially for Vietnamese. In this paper, we bridge this gap through two main contributions:

First, we introduce ViBMR, the first and, to the best of our knowledge, largest multi-sequence brain MRI dataset paired with routine Vietnamese clinical reports. The dataset is collected from a tertiary hospital in Vietnam and comprises 3{,}649 patients with 11{,}875 series. It covers the four MRI sequences commonly used in neuroimaging practice, including T1, T2, FLAIR, and T1c, as well as diverse sequence combinations that reflect the variability of real-world acquisition protocols. The reports are extracted directly from hospital records and exhibit the long-tailed distribution characteristic of clinical practice, ranging from frequent findings such as small vessel disease to rare but clinically significant abnormalities (see Figure~\ref{fig:dataset}). ViBMR therefore provides a large-scale Vietnamese resource for studying medical report generation in a low-resource language setting.

Second, building on ViBMR, we conduct a set of research questions around brain MRI report generation in this low-resource setting. We systematically evaluate the transferability of segmentation-pretrained multi-sequence encoders to report generation, analyze the contribution of individual MRI sequences and their combinations, compare multiple feature aggregation strategies ranging from mean fusion and learnable weighted fusion to spatial gating, channel gating, and cross-attention, and quantify the cross-dataset generalization gap between existing datasets and Vietnamese-origin reports. These experiments establish reliable baselines for ViBMR and provide practical empirical insights for future model design in multi-sequence brain MRI report generation.

\section{Related Work}
\label{sec:related}

\paragraph{Datasets for Brain MRI Report Generation.}


Public brain MRI datasets have substantially advanced medical image analysis, especially for lesion segmentation, disease classification, and algorithmic benchmarking. Representative benchmarks such as BraTS~\cite{Baid2021BraTS}, ISLES~\cite{Hernandez2022ISLES}, the WMH Segmentation Challenge~\cite{Kuijf2019WMH}, and ATLAS~\cite{Liew2018ATLAS} provide high-quality annotations for specific clinical scenarios, including glioma, stroke lesions, white matter hyperintensities, and post-stroke lesion segmentation. However, these resources are primarily designed for anomaly segmentation tasks, rather than for learning the correspondence between multi-sequence brain MRI studies and free-text diagnostic reports. Recent MRI-text resources have begun to address this gap. RadGenome-Brain MRI~\cite{Lei2024AutoRG} provides brain MRI scans, anomaly segmentation masks, and radiologist-written descriptions for abnormal regions, making them suitable for grounded regional report generation. However, RadGenome-Brain MRI is curated from multiple public anomaly segmentation datasets, followed by manually authored findings and impressions for annotated regions~\cite{Lei2024AutoRG}. Therefore, its report structure differs from routine radiology reports directly collected from hospital workflows. MR-RATE~\cite{Forithmus2026MRRATE} further expands MRI-text learning by pairing brain and spine MRI studies with radiology reports and metadata. Another important gap is language coverage. Existing MRI image-report resources remain limited for native low-resource clinical languages. Current brain MRI resources are commonly released or standardized in English, including Chinese-to-English processed reports in AutoRG-Brain~\cite{Lei2024AutoRG} and Turkish-to-English reports in MR-RATE~\cite{Forithmus2026MRRATE}. Beyond brain MRI, Cervical-RG focuses on 3D multi-sequence cervical cancer MRI report generation ~\cite{Zhang2025CervicalRG}, while Auto-Rad and SPINE focus on lumbar or spinal MRI report generation using English or LLM-standardized narrative reports~\cite{Yeasin2024AutoRad,Helmy2026SPINE}. This limitation is important because radiology reports are language, institution, and population-dependent: reporting style, terminology, abbreviation usage, anatomical description granularity, and patient characteristics may differ across hospitals and regions. Therefore, a Vietnamese brain MRI image-report dataset is necessary for studying report generation in a low-resource clinical language setting. In this context, our dataset complements existing brain MRI resources by focusing on routine multi-sequence brain MRI studies paired with Vietnamese clinical reports.

\newcommand{\cmark}{\ding{51}}
\newcommand{\xmark}{\ding{55}}

\begin{table}[h]
\centering
\scriptsize
\setlength{\tabcolsep}{1.2pt}
\renewcommand{\arraystretch}{1.88}
\begin{tabularx}{\columnwidth}{
L{2cm} |
L{1.05cm} |
L{2.1cm} |
C{0.48cm} |
C{0.48cm} |
X
}
\hline
\textbf{Dataset} &
\textbf{Anatomy} &
\textbf{MRI sequences} &
\textbf{Seg.} &
\textbf{Clin.} &
\textbf{Scale} \\
\hline
\hline

MR-RATE~\cite{Forithmus2026MRRATE}
& \makecell[l]{Brain, \\Spine}
& 3D volumetric MRI
& \xmark
& \cmark
& \makecell[l]{83,425 patients /\\705,254 volumes} \\

SSPH Dataset~\cite{Lei2024AutoRG}
& Brain
& T1, T2, FLAIR, DWI
& \cmark
& \xmark
& \makecell[l]{12,535 cases /\\30,793 pairs} \\

BTReport-BraTS~\cite{HerasRivera2026BTReport}
& Brain &T1, T1c, T2, FLAIR &  \cmark & \xmark & 1,470 cases \\

\makecell[l]{RadGenome\\-Brain MRI~\cite{Lei2024AutoRG}}
& Brain
& T1, T1c, T2, FLAIR, DWI, ADC
& \cmark
& \xmark
& \makecell[l]{1,007 cases /\\3,408 pairs} \\


HuskyBrain~\cite{HerasRivera2026BTReport}
& Brain & T1,T1c, T2,  FLAIR
& \cmark
& \xmark

& 184 patients\\


\hline
\hline

\textbf{ViBMR (Ours)}
& \textbf{Brain}
& \textbf{T1,T1c, T2, FLAIR }
& \xmark
& \cmark
& \makecell[l]{\textbf{3,649 patients /}\\\textbf{11,875 series}} \\

\hline
\end{tabularx}

\caption{Comparison with representative MRI datasets related to report generation and 3D MRI understanding. ``Seg.'' indicates whether segmentation annotations are provided. ``Clin.'' indicates whether the dataset provides full-study routine clinical reports directly collected from the clinical workflow.}

\label{tab:dataset_comparison}
\vspace{-0.7cm}
\end{table}

\paragraph{Models for 3D MRI report generation.}
Recent medical vision-language models have extended report generation and visual question answering from 2D radiology images to volumetric medical imaging. General 3D medical MLLMs such as M3D-LaMed~\cite{Bai2024M3D} and Med3DVLM~\cite{Xin2025Med3DVLM} demonstrate the feasibility of aligning 3D medical images with clinical text for tasks including image-text retrieval, VQA, and report generation. For brain MRI, AutoRG-Brain introduces grounded report generation with lesion and anatomical masks~\cite{Lei2024AutoRG}, PIRTA improves the factuality of acute ischemic stroke report generation from 3D DWI MRI using paired image-domain retrieval and text-domain augmentation~\cite{Lee2024PIRTA}, and Brain3D adapts a 2D medical vision encoder into a native 3D encoder for volumetric brain tumor MRI report generation~\cite{Barone2026Brain3D}. These works demonstrate the potential of 3D visual representations for neuroradiological reporting, but their visual input is commonly formulated as a single volumetric scan 3D image rather than routine multi-sequence brain MRI studies that must be interpreted jointly. Beyond brain MRI, Cervical-RG~\cite{Zhang2025CervicalRG} addresses multi-sequence cervical cancer MRI report generation by concatenating aligned 3D MRI sequences at the input level before ViT-based visual encoding.

\vspace{-0.2cm}
\section{Dataset Construction Process}
\label{sec:dataset}

\begin{figure*}
    \centering
    \includegraphics[width=0.98\linewidth]{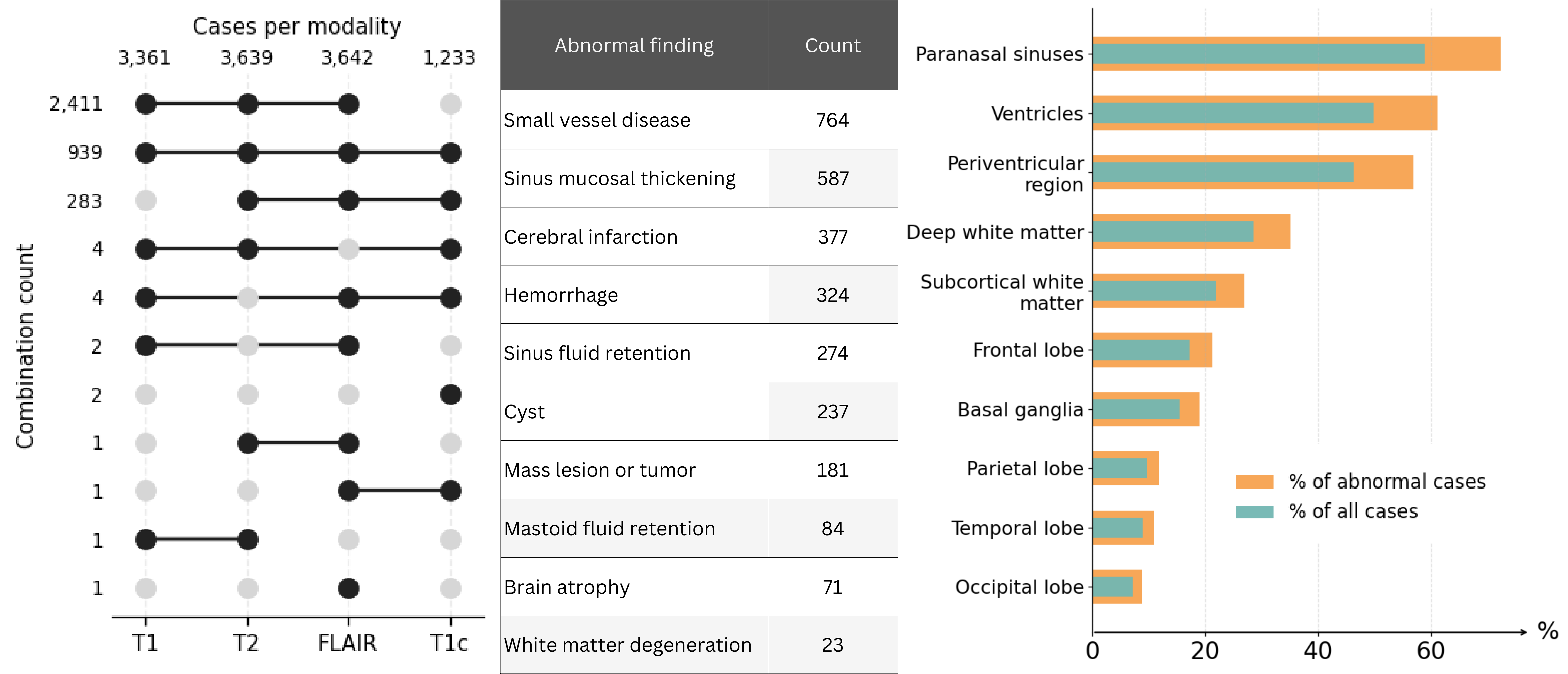}
    \caption{Dataset and distribution analysis of the MRI report dataset. \textbf{(a)} Distribution of MRI sequence availability across the dataset. The bar values above each modality indicate the number of cases containing that sequence, while connected black dots represent specific modality combinations and their corresponding case counts among four MRI sequences: T1-weighted imaging (T1), T2-weighted imaging (T2), Fluid-Attenuated Inversion Recovery (FLAIR), and contrast-enhanced T1-weighted imaging (T1c); \textbf{(b)} Top 10 most frequent abnormal findings; 
\textbf{(c)} Top 10 most frequently mentioned anatomical regions, comparing their prevalence among all cases and abnormal cases.}
    \label{fig:dataset}
    \vspace{-0.5cm}
\end{figure*}

\subsection{Data Preprocessing Pipeline}
We constructed a large-scale real-world brain MRI dataset collected from a major tertiary hospital in Vietnam, consisting of MRI studies from 3,649 patients with diverse intracranial abnormalities and neurological conditions. The radiology reports were extracted from raw hospital PDF documents through a dedicated preprocessing pipeline. To ensure patient privacy and regulatory compliance, all personally identifiable information and sensitive metadata were removed during the anonymization process. We then extracted the findings section of the reports as the ground-truth target for the report generation task. In parallel, we collected the corresponding MRI volumes for each report and aligned them with the extracted report descriptions. Although the number of MRI sequences varies across patients, four modalities are consistently and commonly used in routine neuroimaging practice: T1-weighted (T1), T2-weighted (T2), Fluid-Attenuated Inversion Recovery (FLAIR), and contrast-enhanced T1-weighted imaging (T1c). These modalities provide complementary diagnostic information regarding anatomical structures, lesion characterization, edema, enhancement patterns, and pathological tissue properties. Therefore, we selected these four volumetric MRI sequences as the primary inputs for our evaluation methods. As shown in Fig.~\ref{fig:main_fig}, for each patient, we first organized the DICOM files by study and sequence, and then assembled the ordered slices into volumetric MRI scans using the spatial information stored in the metadata, including slice position, image orientation, and voxel spacing. Since the raw scans were acquired under heterogeneous clinical protocols, the reconstructed volumes may differ in orientation, spatial resolution, field of view, and matrix size. To reduce this variability, all 3D volumes were resampled to an isotropic spatial resolution of $1\times1\times1$ mm. We then applied automated brain extraction using HD-BET~\cite{isensee2019automated} to obtain brain masks, which were used to suppress non-brain tissues and crop the intracranial region. This step removes irrelevant background structures while preserving the clinically meaningful anatomical context, allowing the model to process each sequence as a standardized 3D volume while maintaining anatomical continuity across slices.


\vspace{-0.2cm}
\subsection{Dataset Characteristics}

Our brain MRI dataset contains 3,649 studies 
with heterogeneous sequence availability and diverse clinical findings. Most studies include T1, T2, and FLAIR sequences accounting for 2,411 cases. A smaller subset of 939 studies contains all four sequences, while 283 studies include T2, FLAIR, and T1c. Other modality combinations are rare. 
This distribution indicates that the dataset is dominated by several common acquisition patterns while still reflecting the variability of routine clinical MRI protocols. As shown in Fig.~\ref{fig:dataset}a, this modality imbalance highlights the need for models that can handle incomplete or inconsistent multi-sequence inputs. To characterize the clinical content of the dataset, we extracted structured clinical labels from the radiology reports. We first identified recurrent expressions describing anatomical regions, laterality, abnormal findings, Fazekas grades, and absence of abnormality. A large language model was used to suggest mappings between semantically equivalent expressions, which were subsequently reviewed and verified by a radiologist. Based on the verified mappings, regular-expression rules were constructed to assign each report a multi-label set. 

In addition, the abnormal findings follow a long-tailed distribution (Fig.~\ref{fig:dataset}b). Small vessel disease is the most common abnormality, followed by sinus mucosal thickening, cerebral infarction, hemorrhage, sinus fluid retention, cysts, and mass lesions or tumors. Less frequent findings among the top ten include mastoid fluid retention, brain atrophy, and white matter degeneration. As shown in Fig.~\ref{fig:dataset}c, the anatomical distribution of the most commonly mentioned regions include the paranasal sinuses, ventricles, periventricular region, deep white matter, subcortical white matter, frontal lobe, and basal ganglia. In addition, a single report may contain multiple co-existing findings across different anatomical regions, such as concurrent chronic small vessel disease, sinus abnormalities, infarction, hemorrhage, or mass-related findings. Therefore, the report generation task is not limited to predicting a single dominant abnormality, but requires the model to describe multiple abnormalities and their corresponding anatomical locations within the same study. These statistics suggest that the dataset covers both intracranial abnormalities and extracranial incidental findings, while also exhibiting substantial imbalance across disease categories and anatomical locations. Such characteristics make report generation challenging, as the model must learn from heterogeneous multi-sequence inputs, frequent routine findings, rare but clinically significant abnormalities, and complex anatomical descriptions.

\section{Experiment Setup}
\label{sec:experiments}

\paragraph{Research question.}
Our experiments are designed to answer the following main research questions:


\noindent \textbf{RQ1:} How large is the cross-dataset generalization gap between MR-RATE and ViBMR?
\label{rq:dataset}

\noindent \textbf{RQ2:} Do segmentation-pretrained multi-sequence MRI encoders improve brain MRI report generation?
\label{rq:segment}

\noindent \textbf{RQ3:} Which fusion strategy is most effective for multi-sequence MRI report generation?
\label{rq:fusion}

\noindent \textbf{RQ4:} How do MRI sequences and modality combinations affect report generation?
\label{rq:modality}

\begin{figure*}
    \centering
    \includegraphics[width=\linewidth]{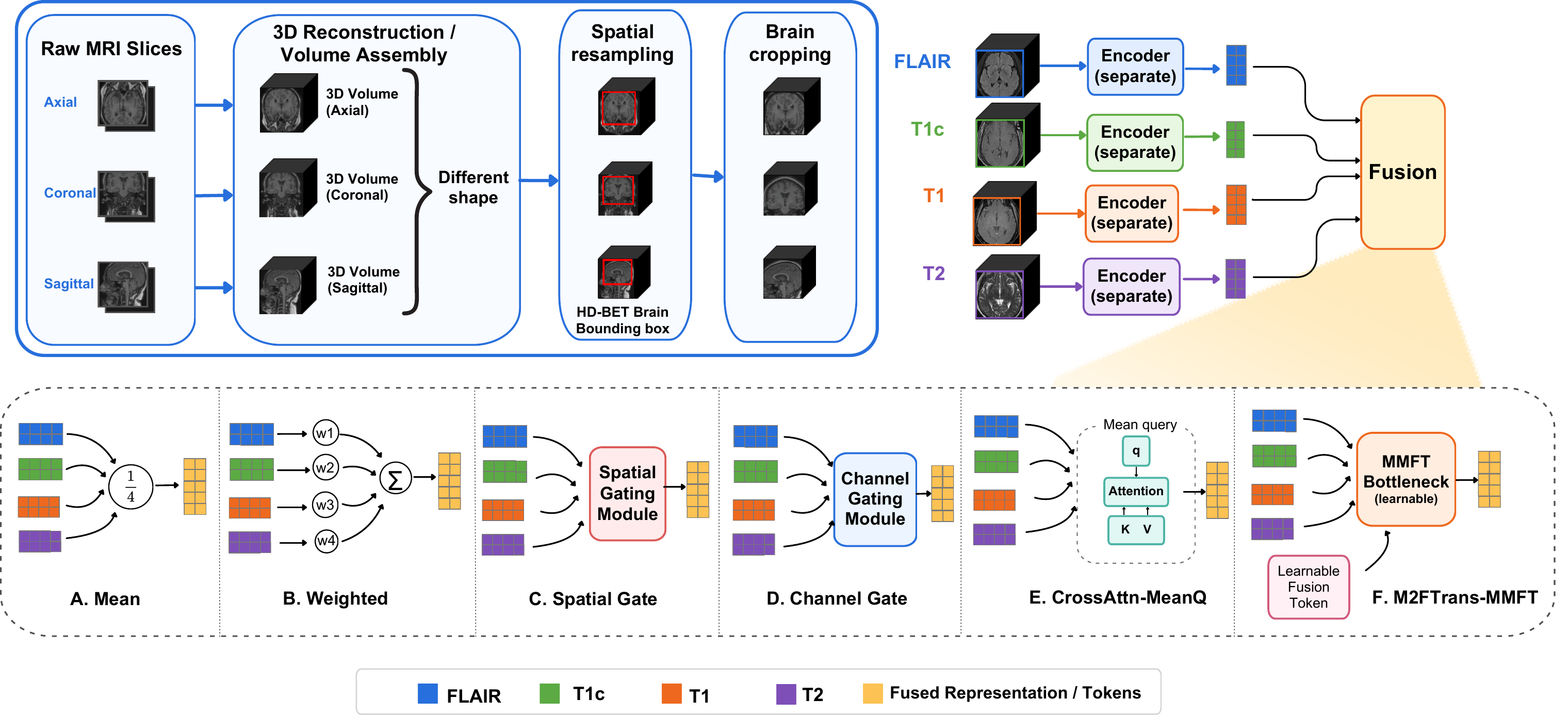}
    \caption{Overview of the proposed MRI report generation pipeline. Raw MRI slices from axial, sagittal, and coronal views are first reconstructed into 3D volumes with different input shapes, followed by voxel-size normalization, brain cropping, and organization into four cropped 3D MRI modalities: FLAIR, T1c, T1, and T2. Each modality is processed by a separate encoder, and the resulting modality-specific visual features are fused before token extraction. The zoom-in panel illustrates the fusion variants evaluated in this study, including mean fusion, learnable weighted fusion, spatial gating, channel gating, mean-query cross-attention, and the M2FTrans-MMFT bottleneck-based fusion strategy.}
    \label{fig:main_fig}
    \vspace{-0.5cm}
\end{figure*}

\vspace{-0.3cm}
\paragraph{Preprocessing and visual feature extraction.}
We use four MRI sequences as input, namely FLAIR, contrast-enhanced T1 (T1c), T1, and T2, as shown in Fig.~\ref{fig:main_fig}. To handle variations in volume size, each study is divided into overlapping 3D patches of size $128 \times 128 \times 128$ with a stride of $64$ along each spatial dimension. If a volume is smaller than the patch size, zero padding is applied before patch extraction. Missing sequences are represented by a modality availability mask and are ignored during modality fusion. For each patch, modality-specific features are extracted by the encoder branches of a segmentation-pretrained M2FTrans~\cite{M2FTrans} or mmFormer~\cite{Zhang2022mmFormer} model and the segmentation decoder branches are removed. Specifically, the FLAIR, T1c, T1, and T2 volumes are passed through corresponding modality encoders, and we take the high-level feature maps before the original bottleneck fusion module. This produces a tensor of shape $[P,4,128,8,8,8]$ for each study, where $P$ denotes the number of extracted patches, $4$ denotes the four MRI sequences, $128$ is the feature dimension, and $8 \times 8 \times 8$ is the spatial feature grid.

\vspace{-0.4cm}
\paragraph{Fusion variants.}
To study the effect of multi-sequence fusion on brain MRI report generation, we compare several fusion strategies over modality-specific feature maps extracted from the encoder branches of segmentation-pretrained models. Let
$\mathbf{X}\in\mathbb{R}^{P\times M\times C\times H\times W\times D}$
denote the extracted feature tensor for one study, where $P$ is the number of 3D patches, $M=4$ is the number of MRI sequences, $C=128$ is the feature dimension, and $H=W=D=8$ is the spatial feature resolution. We define $N=HWD$ and denote the feature vector of patch $p$, modality $m$, and spatial location $s$ by $\mathbf{x}_{p,m,s}\in\mathbb{R}^{C}$, where $s\in\{1,\ldots,N\}$. We use $\mathbf{X}_{p,m}\in\mathbb{R}^{C\times H\times W\times D}$ to denote the corresponding modality-specific feature map. A modality availability mask is denoted by $\mathbf{a}\in\{0,1\}^{M}$, where $a_m=1$ indicates that modality $m$ is available. Figure~\ref{fig:main_fig} provides an overview of the proposed framework. Missing modalities are zero-filled and excluded from masked averaging and softmax operations using $\mathbf{a}_{m}$.


The first three variants correspond to increasingly adaptive weighted aggregation. \textit{Mean fusion} simply averages the available modality feature maps at each patch and spatial location. \textit{Learnable weighted fusion} assigns one trainable global importance score to each MRI sequence, normalizes the scores over available modalities using masked softmax, and shares the resulting weights across patches, spatial locations, and channels. \textit{Spatial gating} further extends this idea by predicting location-dependent modality weights over the 3D feature grid, allowing different anatomical regions to emphasize different MRI sequences.

We further evaluate \textit{channel gating fusion}, which models modality importance at the feature-channel level. Each modality feature map is first globally averaged over the spatial grid to obtain a channel descriptor
$\mathbf{r}_{p,m}=\frac{1}{N}\sum_{s=1}^{N}\mathbf{x}_{p,m,s}\in\mathbb{R}^{C}$.
An MLP $h_{\psi}(\cdot)$ then predicts channel-wise modality logits
$\boldsymbol{\ell}_{p}=h_{\psi}(\mathrm{Concat}_{m=1}^{M}(\mathbf{r}_{p,m}))\in\mathbb{R}^{M\times C}$.
The fused feature is computed as
\begin{equation}
z_{p,c,s}
=
\sum_{m=1}^{M}
\frac{
a_m\exp(\ell_{p,m,c})
}{
\sum_{j=1}^{M} a_j\exp(\ell_{p,j,c})
}
x_{p,m,c,s}.
\end{equation}
Unlike spatial gating, where modality weights vary across spatial locations, channel gating assigns modality weights separately for each feature channel and shares them across all spatial locations within the same patch.

In addition, we test an attention-based variant, denoted as \textit{CrossAttn-MeanQ}. A learnable modality embedding $\mathbf{e}_{m}\in\mathbb{R}^{C}$ is added to each modality token as
$\tilde{\mathbf{x}}_{p,m,s}=\mathbf{x}_{p,m,s}+\mathbf{e}_{m}$.
The query at each spatial location is computed as the masked mean of available embedded modality tokens:
\begin{equation}
\mathbf{q}_{p,s}
=
\frac{
\sum_{m=1}^{M}a_m\tilde{\mathbf{x}}_{p,m,s}
}{
\sum_{m=1}^{M}a_m
}.
\end{equation}
The query sequence is denoted as
$\mathbf{Q}_{p}=[\mathbf{q}_{p,1},\ldots,\mathbf{q}_{p,N}]\in\mathbb{R}^{N\times C}$.
The key and value sequences, $\mathbf{K}_{p},\mathbf{V}_{p}\in\mathbb{R}^{MN\times C}$, are obtained by flattening all embedded modality-specific spatial tokens within the same patch, while tokens from unavailable modalities are masked in the attention operation. The attention-based fused feature is computed as
\begin{equation}
\begin{aligned}
\mathbf{u}_{p,s}
&=
\mathrm{LN}
\left(
\mathbf{q}_{p,s}
+
[
\mathrm{MHA}
(
\mathbf{Q}_{p},\mathbf{K}_{p},\mathbf{V}_{p}
)
]_{s}
\right), \\
\mathbf{z}_{p,s}
&=
\mathbf{u}_{p,s}
+
\mathrm{FFN}(\mathbf{u}_{p,s}).
\end{aligned}
\end{equation}


Here, $[\cdot]_{s}$ denotes the attention output corresponding to spatial location $s$. This design enables token-level cross-modality interaction beyond local weighted summation.

Finally, we include the M2FTrans-MMFT bottleneck fusion as a
segmentation-pretrained fusion baseline. For this baseline, we replace the
lightweight fusion operators with the official pretrained M2FTrans bottleneck
module. For compactness, we denote the ordered modality feature maps of patch
$p$ as $\mathbf{X}_{p}=(\mathbf{X}_{p,1},\ldots,\mathbf{X}_{p,M})$. The M2FTrans bottleneck module is denoted by
$\Phi_{\mathrm{B}}(\cdot)$. It takes $\mathbf{X}_{p}$, the
modality-availability mask, the pretrained fusion-token sequence, and the
pretrained positional embedding as input. We use the updated fusion-token
sequence as the fused cross-modality representation:
\begin{equation}
\begin{aligned}
\mathbf{T}_{p}^{\prime}
=
\pi_{\mathrm{T}}
\Big[
\Phi_{\mathrm{B}}
\big(
\mathbf{X}_{p},
\mathbf{a},
\mathbf{T}_{p}^{0},
\mathbf{P}_{pos};
\boldsymbol{\eta}
\big)
\Big],
\end{aligned}
\end{equation}


where $\pi_{\mathrm{T}}(\cdot)$ selects the updated fusion-token sequence from
the outputs of the bottleneck module. Here,
$\mathbf{T}_{p}^{0}\in\mathbb{R}^{N\times C}$ is the pretrained fusion-token
sequence copied for patch $p$, $\mathbf{P}{pos}$ is the pretrained positional
embedding, and $\boldsymbol{\eta}$ denotes the learnable parameters of the
bottleneck module. During training, the bottleneck parameters, fusion-token
sequence, and positional embedding are fine-tuned. The updated fusion-token sequence is reshaped back to the bottleneck feature grid ($N\rightarrow H\times W\times D$). This baseline evaluates whether a fusion module pretrained for incomplete multi-modal MRI segmentation can provide useful fused representations for report generation.

After each fusion strategy, the fused feature map
$\mathbf{Z}_{p}\in\mathbb{R}^{C\times8\times8\times8}$
is adaptively average pooled to a $2\times2\times2$ grid. Let $\Omega_r$ denote the set of spatial locations assigned to pooled cell $r$, where $r\in\{1,\ldots,8\}$. Each visual token is obtained by average pooling within the corresponding cell:
\begin{equation}
\mathbf{v}_{p,r}
=
\frac{1}{|\Omega_r|}
\sum_{s\in\Omega_r}
\mathbf{z}_{p,s}.
\end{equation}
Since the pooled grid contains $2\times2\times2=8$ spatial cells, each patch produces 8 visual tokens, where each token is a $C$-dimensional feature vector. The study-level visual token sequence is obtained by concatenating tokens from all patches. If the sequence is longer than 64 tokens, it is evenly divided into 64 groups and each group is averaged into one token, producing a compact study-level representation for the LLM.


\vspace{-0.4cm}
\paragraph{Language model.}
The fused visual tokens are mapped into the language-model embedding space using a two-layer visual projector. The projector consists of layer normalization, a linear projection from the visual feature dimension to the LLM hidden size, a GELU activation, and a second linear layer. The projected visual embeddings are concatenated with the textual prompt embeddings and fed into Qwen3-4B-Instruct. The model is trained to generate the findings section of the Vietnamese brain MRI report.

\vspace{-0.4cm}
\paragraph{Implementation details and hyperparameters.}
We split ViBMR into train/validation/test sets with an 8/1/1 ratio at the patient level. Each patient contributes exactly one MRI study, with no patient overlap across splits. During report-generation training, we optimize the fusion module, the visual projector, and the LoRA parameters of the language model. We use LoRA with rank $r=16$, scaling factor $\alpha=32$, and dropout rate $0.05$. All fusion variants are trained under the same optimization setting for a fair comparison. We train for 20 epochs with a batch size of 1 and use gradient accumulation over 8 steps. The maximum number of visual tokens is set to 64, and the maximum target report length is set to 1024 tokens. We use AdamW as the optimizer. The learning rate is set to $1\times10^{-4}$ for both the fusion module and the visual projector, and $1\times10^{-5}$ for LoRA parameters.

\section{Evaluation}
\label{sec:evaluation}

\subsection{Evaluation Metrics}

We evaluate report generation using BLEU-1, BLEU-4, ROUGE-1, ROUGE-2, ROUGE-L, and BERTScore. While BLEU and ROUGE measure lexical overlap, BERTScore evaluates semantic similarity based on contextualized representations.

To further assess clinical correctness, we apply the previously described clinical label extraction method to both the reference and generated reports. The extracted labels are then compared at the sample level to measure agreement in clinically relevant findings, including anatomical region, laterality, abnormal finding, Fazekas grade, and absence of abnormality. For sample $i$, let $T_i$ and $G_i$ denote the clinical label sets extracted from the reference and generated reports, respectively. We define $TP_i = |T_i \cap G_i|$, $FP_i = |G_i \setminus T_i|$, and $FN_i = |T_i \setminus G_i|$. The sample-level clinical F1 score is computed as
\begin{equation}
F1_i = \frac{2TP_i}{2TP_i + FP_i + FN_i}.
\end{equation}

The final Clinical-F1 score is obtained by averaging the sample-level F1 scores over all evaluable reports.

\subsection{Result and Analysis}

\subsubsection*{\hyperref[rq:dataset]{RQ1}. Cross-dataset Generalization.}

\begin{figure}[h]
    \centering
    \includegraphics[width=\linewidth]{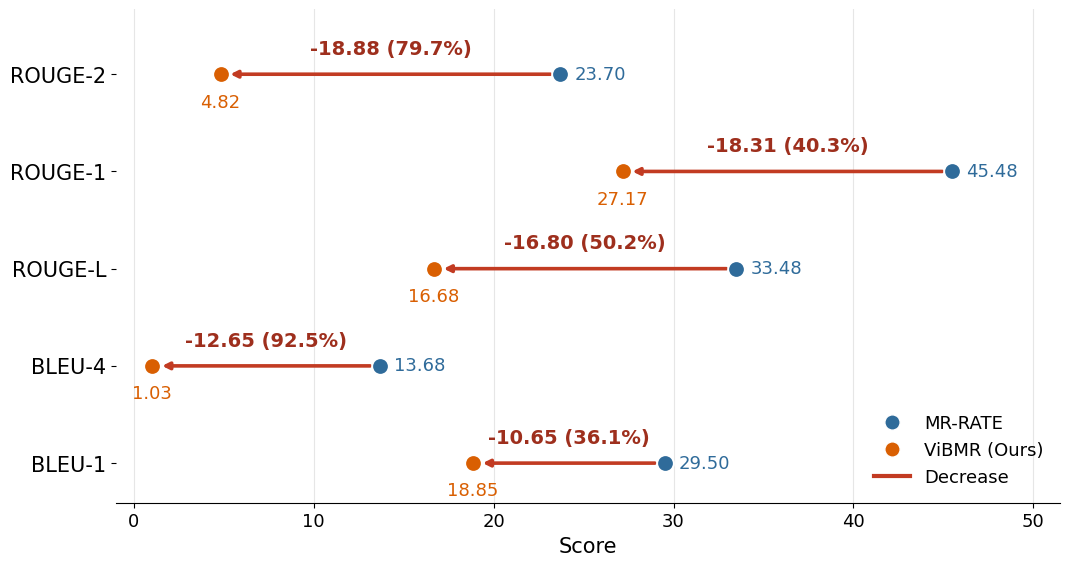}
    \caption{
Cross-dataset generalization after language normalization. The model is trained on MR-RATE and evaluated on both MR-RATE and ViBMR. \emph{Decrease} denotes the performance decrease from MR-RATE to ViBMR. A larger \emph{Decrease} indicates a larger generalization gap.
}
\label{fig:cross_dataset_generalization}
\vspace{-0.4cm}
\end{figure}

Models trained on MR-RATE show limited generalization to Vietnamese-origin reports. To examine this, we use an M2FTrans and Qwen3-4B-Instruct~\cite{qwen3} baseline, where the segmentation-pretrained M2FTrans bottleneck is fine-tuned jointly with LoRA adapters in the LLM. We randomly sample 899 brain MRI cases from MR-RATE and split them into training, validation, and test sets with an 8:1:1 ratio. The model is trained on MR-RATE, then evaluated on both the held-out MR-RATE test split and the ViBMR test set. For ViBMR, reports are translated into English using the GPT-5.4 API model~\cite{openai2026gpt54}, while preserving medical keywords and domain-specific terminology. As shown in Figure~\ref{fig:cross_dataset_generalization}, performance of the model drops from 29.50 to 18.85 BLEU-1, 13.68 to 1.03 BLEU-4, and 33.48 to 16.68 ROUGE-L when moving from MR-RATE to ViBMR test set. The largest gaps occur in ROUGE-1 and ROUGE-2, both exceeding 18 points, indicating a substantial cross-dataset performance gap. However, this experiment does not isolate the specific source of this degradation. Possible contributing factors include translation artifacts, differences in reporting style and clinical expression, population and disease distributions, scanner and imaging protocol differences, and overall dataset composition. Therefore, the observed gap should be interpreted as evidence of cross-dataset domain shift rather than being attributed to any single factor. These results suggest that ViBMR provides a complementary setting for studying the robustness and cross-dataset generalization of medical vision-language models in a low-resource clinical and linguistic context.

\subsubsection*{\hyperref[rq:segment]{RQ2}. Effect of Segmentation-Pretrained MRI Encoders.}

\begin{table}[b]
\vspace{-0.3cm}
\centering
\scriptsize
\setlength{\tabcolsep}{2.2pt}
\renewcommand{\arraystretch}{1.12}
\begin{tabularx}{\columnwidth}{
L{0.17\columnwidth} |
C{0.18\columnwidth}
C{0.18\columnwidth}
C{0.22\columnwidth}
C{0.15\columnwidth}
}
\hline
&
\makecell[c]{\textbf{M2FTrans}\\\textbf{(Encoder)}} &
\makecell[c]{\textbf{mmFormer}\\\textbf{(Encoder)}} &
\textbf{AutoRG-Brain} &
\textbf{M3D} \\
\hline
BLEU-1  & \textbf{68.09} & \underline{64.31} & 54.44 & 42.09 \\
BLEU-4  & \textbf{60.04} & \underline{58.22} & 47.25 & 35.32 \\
ROUGE-1 & \underline{75.02} & 72.59 & \textbf{78.12} & 57.06 \\
ROUGE-2 & \textbf{66.67} & \underline{65.68} & 62.80 & 47.52 \\
ROUGE-L & \textbf{70.17} & \underline{68.69} & 65.85 & 50.94 \\
BERT & \textbf{87.02} & \underline{85.55} & 83.30 & 80.80 \\
CLIN & \textbf{69.47} & \underline{66.18} & 58.79 & 45.28 \\
\hline
\end{tabularx}
\caption{Comparison of brain MRI report generation model performance under the same LoRA adapter setting for decoder models. M2FTrans and mmFormer are used as encoder baselines with Qwen3-4B-Instruct as the language model. BERT denotes the BERTscore F1, while CLIN denotes the Clinical-F1 score. The best value for each metric is shown in \textbf{bold}, and the second-best value is \underline{underlined}.}
\label{tab:model_comparison}
\end{table}

Segmentation-pretrained multi-sequence MRI encoders provide stronger visual representations for brain MRI report generation than the evaluated single-modality baselines. Using the default setting of the encoder baselines, under the same LoRA adapter setting on the decoder-side, the M2FTrans encoder with Qwen3-4B-Instruct achieves the best performance in six of seven metrics, as shown in Table~\ref{tab:model_comparison}. Moreover, mmFormer encoder baseline also obtains competitive results and ranks second on most metrics, indicating that segmentation-pretrained MRI encoders can be effectively transferred from dense prediction tasks to clinical text generation. The comparison further suggests that multi-sequence input is important for generating more accurate brain MRI reports. M2FTrans and mmFormer baselines can process multiple MRI sequences simultaneously, whereas AutoRG-Brain and M3D are evaluated using single-modality input. Their stronger overall performance also indicates that complementary information from different MRI sequences provides richer visual evidence for report generation. Although AutoRG-Brain obtains the highest ROUGE-1 score, its lower scores across the remaining metrics suggest that this advantage is mainly limited to unigram-level overlap and does not consistently extend to longer phrase-level, semantic, or clinical agreement. These results answer \hyperref[rq:segment]{RQ2} by showing that segmentation-pretrained multi-sequence MRI encoders are effective visual backbones for brain MRI report generation.

\subsubsection*{\hyperref[rq:fusion]{RQ3}. Effect of Different Fusion Strategies.}

For M2FTrans, mean fusion achieves the highest BLEU-4 and Clinical-F1, while channel gating leads in ROUGE-2 and BERTScore. For mmFormer, weighted fusion performs best on BLEU-4, ROUGE-2, and BERTScore, whereas mean fusion leads in Clinical-F1 (Table~\ref{tab:fusion_results}). Overall, lightweight fusion methods generally outperform CrossAttn-MeanQ, suggesting that segmentation-pretrained features can be effectively aggregated without complex attention mechanisms. However, the optimal fusion strategy varies across encoders and evaluation metrics.


\begin{table}[h]
\centering
\scriptsize
\setlength{\tabcolsep}{2.4pt}
\renewcommand{\arraystretch}{1.12}
\begin{tabularx}{\columnwidth}{
L{0.16\columnwidth} |
L{0.2\columnwidth} |
C{0.12\columnwidth}
C{0.14\columnwidth}
C{0.14\columnwidth}
C{0.14\columnwidth}
}
\hline
\textbf{Encoder} &
\textbf{Fusion} &
\textbf{BLEU-4} &
\textbf{ROUGE-2} &
\textbf{BERT} &
\textbf{CLIN} \\
\hline

\multirow{6}{*}{\textbf{M2FTrans}}
& mean              & \textbf{59.13} & \underline{72.62} & 87.25 & \textbf{72.33} \\
& weighted          & 51.02 & 66.75 & 85.48 & 68.56 \\
& spatial\_gate     & 58.32 & 72.24 & \underline{87.25} & \underline{71.62} \\
& channel\_gate     & \underline{58.95} & \textbf{72.64} & \textbf{87.37} & 70.41 \\
& crossattn\_meanq  & 53.8  & 68.95 & 85.89 & 69.84 \\
& \textit{m2ftrans\_mmft} & \textit{58.3} & \textit{72.22} & \textit{87.10} & \textit{70.84} \\
\hline

\multirow{6}{*}{\textbf{mmFormer}}
& mean              & 59.79 & 72.94 & 87.44 & \textbf{71.75} \\
& weighted          & \textbf{60.8} & \textbf{73.45} & \textbf{87.49} & \underline{71.62} \\
& spatial\_gate     & 59.51 & 73.08 & \underline{87.45} & 71.40 \\
& channel\_gate     & \underline{60.39} & \underline{73.42} & 87.37 & 71.36 \\
& crossattn\_meanq  & 50.3  & 66.7  & 85.22 & 68.89 \\
& \textit{mmformer} & \textit{55.82} & \textit{70.31} & \textit{86.45} & \textit{70.55} \\
\hline
\end{tabularx}

\caption{Comparison of different fusion strategies for MRI report generation using visual encoders initialized from M2FTrans and mmFormer checkpoints. The best value in each encoder group is shown in \textbf{bold}, and the second-best value is \underline{underlined}. The final row in each group denotes the original baseline configuration.}
\label{tab:fusion_results}
\vspace{-0.5cm}
\end{table}

\subsubsection*{\hyperref[rq:modality]{RQ4}. Contribution of Different MRI Sequences.}

For this modality ablation study, we select the two best-performing fusion strategies and two original baselines, evaluating selected modality combinations on the same subset of 939 studies containing all four sequences. Different combinations are constructed by masking selected sequences. As shown in Table~\ref{tab:modality_ablation}, most methods benefit from multi-sequence inputs, with the strongest performance generally obtained using T1+T2+FLAIR or all four modalities. However, improvements vary across fusion strategies. In particular, M2FTrans-MMFT achieves relatively consistent performance across modality combinations, potentially benefiting from its pretrained fusion tokens that capture modality relationships. Overall, these results highlight the benefits of complementary MRI sequences, the effectiveness of T1+T2+FLAIR, and the importance of fusion design for exploiting multi-modality information.


\begin{table}[t]
\centering
\scriptsize
\setlength{\tabcolsep}{1.8pt}
\renewcommand{\arraystretch}{1.12}
\begin{tabularx}{\columnwidth}{
L{0.14\columnwidth} |
L{0.18\columnwidth} |
L{0.18\columnwidth} |
C{0.10\columnwidth}
C{0.10\columnwidth}
C{0.10\columnwidth}
C{0.10\columnwidth}
}
\hline
\textbf{Encoder} &
\textbf{Method} &
\textbf{Modality} &
\textbf{B-4} &
\textbf{R-L} &
\textbf{BERT} &
\textbf{CLIN} \\
\hline

\multirow{12}{*}{\textbf{M2FTrans}}
& \multirow{6}{*}{mean}
& T1              & 40.70 & 58.99 & 80.25 & 50.09 \\
&
& T2              & 40.94 & 61.46 & 82.57 & 62.79 \\
&
& FLAIR           & 48.19 & 66.62 & 83.66 & 61.83 \\
&
& T1+T2           & 50.53 & 67.52 & 84.43 & 62.57 \\
&
& T1+T2+FLAIR     & \textbf{59.29} & \textbf{74.55}
                    & \textbf{87.21} & \textbf{71.75} \\
&
& All             & \underline{53.77} & \underline{70.58}
                    & \underline{85.15} & \underline{69.71} \\
\cmidrule(lr){2-7}

& \multirow{6}{*}{m2ftrans\_mmft}
& T1              & 54.11 & 70.79 & 85.23 & 65.63 \\
&
& T2              & \underline{56.22} & \underline{72.51}
                    & \underline{85.92} & 67.62 \\
&
& FLAIR           & \textbf{56.67} & \textbf{72.85}
                    & 85.44 & 66.24 \\
&
& T1+T2           & 54.55 & 71.03 & 85.52 & 67.70 \\
&
& T1+T2+FLAIR     & 55.72 & 71.99
                    & \textbf{86.18} & \textbf{69.82} \\
&
& All             & 55.11 & 71.68
                    & 85.43 & \underline{68.41} \\
\hline

\multirow{12}{*}{\textbf{mmFormer}}
& \multirow{6}{*}{mmformer}
& T1              & 39.44 & 60.38 & 82.09 & 54.99 \\
&
& T2              & 55.16 & 71.28 & 85.01 & 65.10 \\
&
& FLAIR           & 23.62 & 50.49 & 78.89 & 58.70 \\
&
& T1+T2           & 52.29 & 69.09 & 84.19 & 62.90 \\
&
& T1+T2+FLAIR     & \textbf{58.54} & \textbf{74.58}
                    & \textbf{86.95} & \textbf{71.41} \\
&
& All             & \underline{56.11} & \underline{72.30}
                    & \underline{85.47} & \underline{67.27} \\
\cmidrule(lr){2-7}

& \multirow{6}{*}{weighted}
& T1              & 42.01 & 60.06 & 81.23 & 53.55 \\
&
& T2              & 49.24 & 66.95 & 83.81 & 64.25 \\
&
& FLAIR           & 54.09 & 70.89
                    & \underline{85.71} & \underline{67.75} \\
&
& T1+T2           & 46.40 & 63.96 & 82.20 & 59.89 \\
&
& T1+T2+FLAIR     & \textbf{58.52} & \textbf{73.83}
                    & \textbf{86.89} & \textbf{70.57} \\
&
& All             & \underline{55.51} & \underline{71.15}
                    & 85.43 & 67.24 \\
\hline
\end{tabularx}

\caption{Comparison of modality importance across different fusion methods using M2FTrans and mmFormer encoder checkpoints. In the Modality column, the ''+'' symbol denotes combining additional sequences. ''All'' denotes using all four available MRI modalities. B-4 and R-L denote BLEU-4 and ROUGE-L, respectively. The best value for each metric within each method is shown in \textbf{bold}, and the second-best value is \underline{underlined}.}
\label{tab:modality_ablation}
\vspace{-0.5cm}
\end{table}  
\section{Conclusion}
\label{sec:conclusion}

In this paper, we introduced ViBMR, a large-scale multi-sequence brain MRI dataset paired with routine Vietnamese clinical reports, and presented a systematic study of automated brain MRI report generation. Our experiments show that segmentation-pretrained multi-sequence MRI encoders provide effective visual representations, while lightweight fusion strategies such as mean fusion, weighted fusion, and gating are generally more stable than complex attention-based fusion. We also observe that multi-sequence inputs improve report generation and that models trained on existing English MRI-report datasets exhibit a clear generalization gap when evaluated on Vietnamese-origin reports. These findings highlight the importance of clinically grounded low-resource datasets and robust multi-sequence fusion for future brain MRI report generation systems.

\section*{Ethical Statement}
This retrospective study was approved by the Institutional Review Board (IRB), with informed consent waived due to minimal risk. All imaging data and reports were de-identified before research use, with personally identifiable information (PII) removed. The dataset will be released for non-commercial research under specified usage terms.
\section*{Acknowledgements}
This research is funded by Vietnam National Foundation for Science and Technology Development (NAFOSTED) under grant number 24/2025/TN.

\section*{Contribution Statement}
Phi Le Nguyen is the corresponding author.

\newpage
{
    \small
    \bibliographystyle{ieeenat_fullname}
    \bibliography{main}
}

\end{document}